\documentclass[conference,usletter]{IEEEtran}
\usepackage{times,amsmath,amssymb}
\usepackage{slashbox}
\usepackage{graphicx}
\usepackage{xcolor}
\usepackage{algorithm2e}
\usepackage{algorithmicx}
\usepackage{setspace}
\usepackage[export]{adjustbox}
\usepackage{multirow}
\usepackage[noend]{algpseudocode}
\usepackage{array,multirow}
\usepackage{makecell}
\usepackage{mathtools}
\usepackage{soul}

\begin{document}

\title{Remote Multilinear Compressive Learning with Adaptive Compression}
\author{\IEEEauthorblockN{Dat Thanh Tran\IEEEauthorrefmark{1}, Moncef Gabbouj\IEEEauthorrefmark{1}, Alexandros Iosifidis\IEEEauthorrefmark{2}}
\IEEEauthorblockA{\IEEEauthorrefmark{1}Department of Computing Sciences, Tampere University, Tampere, Finland\\
\IEEEauthorrefmark{2}Department of Electrical and Computer Engineering, Aarhus University, Aarhus, Denmark\\
Email:\{thanh.tran,moncef.gabbouj\}@tuni.fi, ai@ece.au.dk}\\

}

\maketitle

\begin{abstract}
Multilinear Compressive Learning (MCL) is an efficient signal acquisition and learning paradigm for multidimensional signals. The level of signal compression affects the detection or classification performance of a MCL model, with higher compression rates often associated with lower inference accuracy. However, higher compression rates are more amenable to a wider range of applications, especially those that require low operating bandwidth and minimal energy consumption such as Internet-of-Things (IoT) applications. Many communication protocols provide support for adaptive data transmission to maximize the throughput and minimize energy consumption. By developing compressive sensing and learning models that can operate with an adaptive compression rate, we can maximize the informational content throughput of the whole application. In this paper, we propose a novel optimization scheme that enables such a feature for MCL models. Our proposal enables practical implementation of adaptive compressive signal acquisition and inference systems. Experimental results demonstrated that the proposed approach can significantly reduce the amount of computations required during the training phase of remote learning systems but also improve the informational content throughput via adaptive-rate sensing. 
\end{abstract}


\section{Introduction}\label{S:Intro}

Internet-of-Things (IoT) technologies  allow the deployment of smart sensors in every corner of the world, including the most remote areas. This is thanks to the development of low-cost electronics and embedded devices, as well as the advancements in wireless communication technologies such as those in Low Power Wide Area Networks (LPWAN). An IoT system is a cyber-physical system in which the physical (client) side often consists of a network of smart sensors and embedded devices deployed in various physical locations, with capability to acquire signals and perform light-weight computation before transmitting them to a network cloud. On the cyber (server) side, the collected data, which is aggregated and analyzed by the network server, can be used for automatic decision making and intelligent services \cite{qi2017dnn, mohammadi2018deep, zhou2019edge}. Since the whole idea of IoT systems is built on information gathering, efficient signal acquisition and transmission play an important role in IoT applications. Among different factors, energy consumption and computational complexity are key factors that determine the efficiency of data collection in IoT services since deployed sensors are often small, low-power embedded devices having small computational capacity and limited energy source.     

Under such system requirements, Compressive Sensing \cite{candes2008introduction}, which is an efficient signal acquisition technology, is a suitable solution for sensor data collection in an IoT stack. The efficiency in CS devices comes from the fact that signals are sampled, quantized and compressed at the same time, at the hardware level, i.e., on the sensors. This is different from traditional sensors from which we obtain a large amount of discrete samples or raw measurements from the signal, and the compression step is conducted after the acquisition step, at the software level. In CS, the signal is measured and compressed at the same time, before being registered in the memory during the sampling phase. Due of this, a CS sensor requires a much smaller memory size to store temporary data, and outputs a discrete signal with a significantly lower number of measurements for storage and transmission. 

Given an input signal $\mathbf{y} \in \mathbb{R}^{I}$ that has been sampled and quantized, a CS device, instead of outputing $\mathbf{y}$, it registers a compressed version of $\mathbf{y}$ by linearly projecting $\mathbf{y}$ onto a lower-dimensional space as follows: 
\begin{equation}\label{eq1}
	\mathbf{z} = \mathbf{\Phi} \mathbf{y}
\end{equation}
where $\mathbf{z} \in \mathbb{R}^{M}$ denotes the measurement of the input signal and $\mathbf{\Phi} \in \mathbb{R}^{M\times I}$ denotes the projection matrix. The number of measurements obtained by a CS device is often significantly lower than the dimension of $\mathbf{y}$, i.e., $M \ll I$, so $\mathbf{\Phi} \in \mathbb{R}^{M \times I}$ is a fat matrix. $\Phi$ is also referred to as the sensing operator or sensing matrix.  

Here we should note that what we obtain from a traditional sensor is $\mathbf{y}$, which is often sampled at a sampling rate higher than the Nyquist rate to ensure perfect reconstruction of the input signal. Since the dimension of $\mathbf{z}$ is lower than that of $\mathbf{y}$, the signal is undersampled in a CS device. Although  Shanon Theorem on ideal sampling specifies that a signal must be sampled at a higher rate than the Nyquist rate to guarantee perfect reconstruction, the undersampled signal registered by a CS device can still be recovered near perfectly if the input signal can be expressed with a sparse representation in some domain and the sensing matrix $\Phi$ possesses certain properties \cite{candes2006stable, donoho2006compressed}. These results are known as the CS theory, which is the foundation and motivation for the developments of compressive sensing and compressive learning methods.       

Although signals can be acquired at a very low cost using the CS paradigm, reversing them to the original domains is a daunting task since it involves solving an optimization problem to determine the sparse representation and the corresponding bases of the signal. In some use cases, signal recovery is necessary and the quality of signal reconstruction plays an important role. For example in medical imaging for health  diagnosis, the higher the resolution and fidelity of the reconstructed signal are, the better chance of distinguishing between health abnormality and noise generated from the sensing process. However, for many applications, the purpose of acquiring signals is to perform classification or regression of some kind, rather than reconstructing the original signals. For example, in forest fire detection and monitoring via unmanned aerial vehicles, the objective of forest imaging is to automatically detect and locate potential locations of fire. In some applications that involve humans such as those in smart buildings, the reconstruction of data should be avoided since this step can potentially disclose private information. In fact, the majority of IoT services and automation systems gather data mainly for intelligence and decision purposes, rather than high-fidelity reconstruction.          

Due of the aforementioned reasons, researchers have proposed machine learning models that are tailored to work directly with the compressed measurements obtained from CS devices without going through a proxy signal recovery step. Methodologies developed under this objective form a research topic known as Compressive Learning (CL). The early works in CL took a similar approach to CS literature, relying on random valued sensing operators and focusing on theoretical guarantees for models operating on compressed measurements \cite{calderbank2012finding, davenport2007smashed, davenport2010signal, reboredo2013compressive}. The reliance on random sensing matrices not only exempts us from the problem of joint estimation of the parameters of the learning model and the sensing operator but also allows us to take advantage of existing theoretical results on random linear projections \cite{baraniuk2009random}. Besides, theoretical results on perfect signal reconstruction in CS were derived from random sensing matrices. Since the ability for high-fidelity signal recovery also implies the preservation of signal content in the compressed measurement, there is an assurance that the machine learning model is estimated with the same degree of informational content in such cases.      

With wide adoption of modern stochastic optimization methods during the past decade, more recent works in CL have switched from using random sensing matrices to optimized ones, which are jointly estimated with the model's parameters. Although theoretical results for the generalization ability of  compressive learning models following the end-to-end learning approach are yet to be derived, several works demonstrated its superior performance over prior setups that use random sensing operators \cite{adler2016compressed, lohit2016direct, hollis2018compressed, zisselman2018compressed, tran2019multilinear, tran2020multilinear}. Among these end-to-end compressive learning models, Multilinear Compressive Learning (MCL) \cite{tran2019multilinear} is the leading solution in both inference performance as well as computational efficiency for multidimensional signals. This stems from the fact that MCL employs sensing and feature extraction modules that are designed to operate on tensors using multilinear operations. Since a multidimensional input signal is linearly projected along each tensor mode in MCL, the number of computations used to compress a given signal is significantly lower than the projection in Eq. (\ref{eq1}) while the multidimensional structure of the input signal is still retained after the projection. 

Regardless of the approach followed to project the input signals, existing CL formulations require training separate model configurations for different compression rates. This usually results in long experimentation processes in order to determine a suitable trade-off between the compression rate and the inference performance when deploying a CL system. For applications that employ remote compressive sensing, a fixed compression rate for signal acquisition is undesirable. This is because the majority of modern communication standards support adaptive transmission rates to maximize throughput and minimize energy consumption as the transmission environment changes. For example, the Adaptive Data Rate (ADR) scheme is a key feature of the Long-Range Wide Area Networks (LoRaWAN) in IoT technology. By improving CL solutions giving them the ability to acquire the signal at an adaptive compression rate on the sensor level, hence allowing for adaptive degree of signal fidelity, which can be adjusted according to the network status, we can further maximize the signal content throughput of a CL system.    

In this paper, we propose a novel optimization scheme and deployment setup for MCL that enables training multiple model configurations with different compression rates in one shot. The resulting system can be used to evaluate and benchmark different compressed measurement shapes, a process that reduces significantly the experimentation efforts required to find the optimal trade-off between data compression rate and inference performance. In addition, using the proposed optimization scheme, we can obtain sensing operators that produce highly structured compressed measurements which allow us to implement a CS device that is capable of signal acquisition at an adaptive compression rate, thereby solving the aforementioned shortcoming of existing approaches.  

The paper is organized as follows. Section \ref{related-works} reviews the MCL model and related literature. Section \ref{method} provides details of our proposed optimization scheme and deployment setup. In Section \ref{experiments}, we provide detailed information about our experimental protocol, and qualitative and quantitative analyses of the empirical results. Section \ref{conclusion} concludes our work with remarks on the implications of the proposed contributions.

\section{Related Work}\label{related-works}

\begin{figure*}
	\centering
	\includegraphics[width=0.9\textwidth]{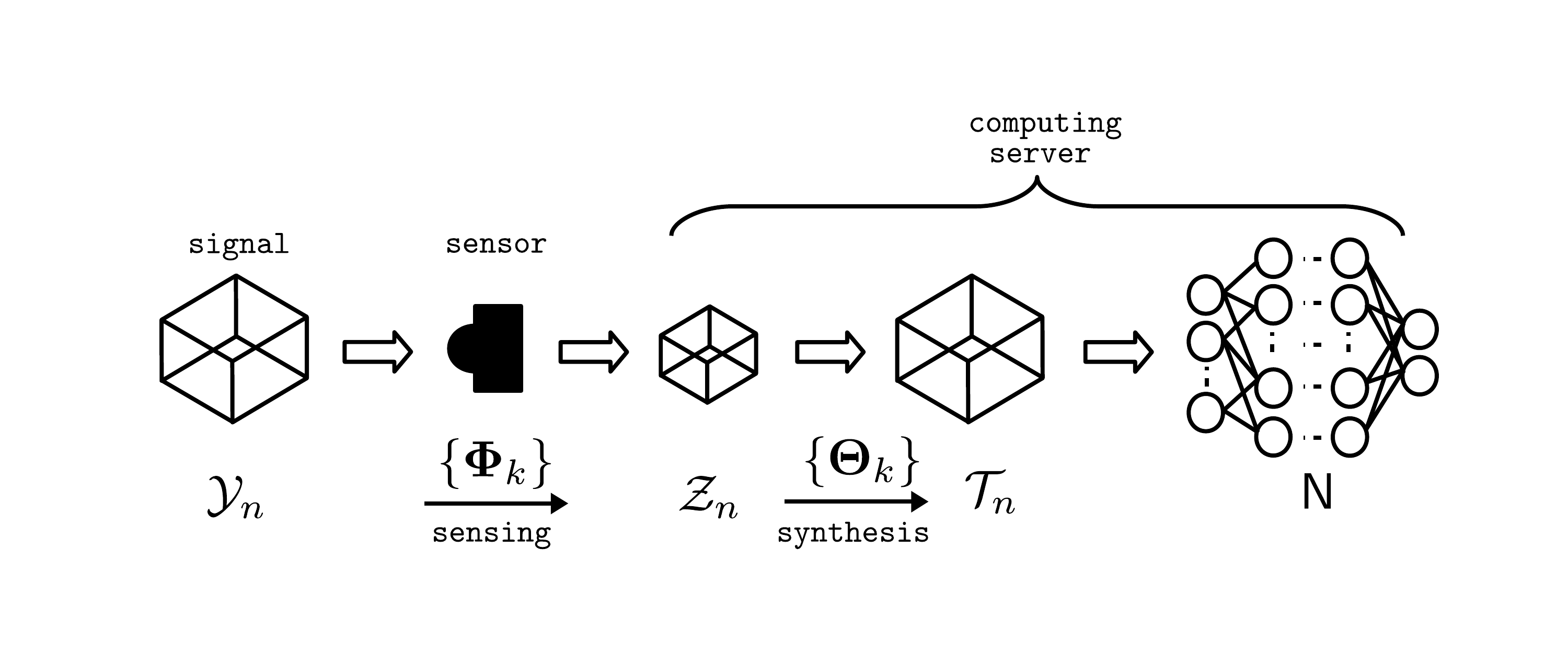}
	\caption{Illustration of the MCL model}\label{f3}
\end{figure*}

In this Section, before the discussion of related works, we will describe the Multilinear Compressive Learning framework, which is the basis of the proposed method. A Multilinear Compressive Learning (MCL) model \cite{tran2019multilinear}, illustrated in Figure \ref{f3}, comprises  three elements: the Multilinear Compressive Sensing (MCS) module, the Feature Synthesis (FS) module and the task-specific neural network $\mathsf{N}$. 

Different from the sensing model described in Eq. (\ref{eq1}), which only considers input signals of vector form, the Multilinear Compressive Sensing (MCS) model employed in MCL performs linear sensing along each of the dimensions of a given multidimensional input signal. Thus, the sensing is executed via a set of sensing matrices, also known as separable sensing operators. More specifically, let us denote the discretized input signal and the compressed measurement obtained from the MCS module as $\mathcal{Y} \in \mathbb{R}^{I_1 \times \dots \times I_K}$ and $\mathcal{Z} \in \mathbb{R}^{M_1 \times \dots \times M_K}$, respectively. In this case, $I_1 \times \dots \times I_K$ represents the signal resolution that the sensor initially samples and quantizes. The MCS compression model is described by the following equation:
\begin{equation}\label{eq2}
\mathcal{Z} = \mathcal{Y} \times_1 \mathbf{\Phi}_1 \times \dots \times_K \mathbf{\Phi}_K,
\end{equation} 
where $\{\mathbf{\Phi}_k \in \mathbb{R}^{M_k \times I_k} \, | \, k=1, \dots, K\}$ denotes the set of separable sensing operators and $\times_k$ denotes the mode-$k$ product between a tensor and a matrix. Detailed description of the mode-$k$ product can be found in \cite{kolda2009tensor}. Basically, this operation linearly transforms every $k$-th dimensional slice of the tensor $\mathcal{Y}$ using the corresponding  matrix $\mathbf{\Phi}_k$.  

Here we should note that at deployment the compressive sensing step described in the above equation (as well as the one in Eq. (\ref{eq1})) is implemented at the hardware level, i.e. on the sensing device. What we obtain from this device is $\mathcal{Z}$ (or $\mathbf{z}$, respectively), i.e. the compressed measurement. Thus, the sensing step in Eq. (\ref{eq2}) should not be viewed as a feature extraction step since they are inherent in the signal acquisition procedure of the CS device. For end-to-end CL approaches, during development and optimization, we often simulate this signal acquisition step at the software level, using the high resolution signal $\mathcal{Y}$ that was sampled above the Nyquist rate using a standard sensor in order to optimize the sensing operators.     

Given the compressed measurement $\mathcal{Z}$, MCL synthesizes a high-dimensional tensor feature that is relevant for the learning task by the FS module. In order to preserve the multidimensional structure of the compressed measurement, the FS module proposed in \cite{tran2019multilinear} also employs a multilinear transform. However, since the FS module is implemented at the software level, usually on the computing cloud for remote sensing applications, the FS module is not constrained to the use of multilinear operations as long as the tensor structure of the signal is preserved. For example, the authors in \cite{tran2020multilinear} extended the original design of the MCL model in \cite{tran2019multilinear} with a FS module that contains several convolution and up-sampling layers. The choice of the FS module mainly depends on the operating power of the computing server. In this work, we investigate our optimization scheme using the original design in \cite{tran2019multilinear} for the FS component, which can be described by the following equation:
\begin{equation}\label{eq3}
\mathcal{T} = f_{\mathsf{FS}}(\mathcal{Z}) = \mathcal{Z} \times_1 \mathbf{\Theta}_1 \times \dots \times_K \mathbf{\Theta}_K,
\end{equation} 
where $\mathcal{T} \in \mathbb{R}^{I_1 \times \dots \times I_K}$ denotes the synthesized feature, $\{\mathbf{\Theta}_k \in \mathbb{R}^{I_k \times M_k} \, | \, k=1, \dots, K\}$ denotes the parameters of the FS component, and $f_{\mathsf{FS}}$ denotes its functional form. 
	
Finally, $\mathcal{T}$ is introduced to a task-specific neural network $\mathsf{N}$ to produce a prediction for the given compressed measurement. The architecture of $\mathsf{N}$, which depends on the given problem at hand, is the same architecture that one would use to classify uncompressed signals, i.e., the high-resolution signal $\mathcal{Y}$. For this reason, the dimensions of $\mathcal{T}$ in Eq. (\ref{eq3}) are the same as those of $\mathcal{Y}$, i.e. the high-resolution signal.

In MCL, all model parameters are jointly optimized in an end-to-end manner by a stochastic gradient descent optimizer. Different from the conventional approach where the parameters are initialized with random values, MCL determines the initial values of each component's parameters by solving two optimization problems: one for the MCS and FS modules and the other for the task-specific neural network. As we mentioned previously, for methods that optimize the sensing operator we need a labeled set of high-resolution signals, which is obtained using a standard sensor sensing at higher-than-Nyquist rate, in order to simulate the compressive sensing step during optimization. Let us denote $N$ as the number of samples in the training set. In addition, we also denote the $n$-th high-resolution sample in the training set as $\mathcal{Y}_n$, its compressed measurement as $\mathcal{Z}_n$, the corresponding synthesized feature as $\mathcal{T}_n$ and the label as $\mathbf{c}_n$. The initial parameters' values of the MCS and FS modules are obtained by solving Higher Order Singular Value Decomposition (HOSVD) \cite{de2000multilinear} using the set of high-resolution signals $\{\mathcal{Y}_n \, | \, n=1, \dots, N\}$. More specifically, let $\mathcal{Y} \in \mathbb{R}^{I_1 \times \dots \times I_K \times N}$ denotes the concatenation of all $N$ samples along the $(K+1)$-th dimension. In addition, let us denote the HOSVD of $\mathcal{Y}$ along the first $K$ dimensions as follows:
\begin{equation}\label{eq4}
\mathcal{Y} = \mathcal{S} \times_1 \mathbf{U}_1 \times \dots \times_K \mathbf{U}_K
\end{equation}
The sensing operators are initialized by setting $\mathbf{\Phi}_k = \mathbf{U}_k^{T}$, and the FS parameters are initialized by setting $\mathbf{\Theta}_k = \mathbf{U}_k$. Here $\mathcal{S} \in \mathbb{R}^{M_1 \times \dots \times M_K \times N}$ denotes the tensor core that contains the singular values and $\mathbf{U}_k \in \mathbb{R}^{M_k \times I_k}$ ($k=1, \dots, K$) denotes the factor matrices of the decomposition. By following the above initialization scheme for the sensing and the feature extraction matrices, the energy of $\mathcal{Y}_n$ in $\mathcal{T}_n$ is preserved since $\mathbf{U}_k$ ($k=1, \dots, K$) are unitary matrices.  

The parameters of the task-specific neural network $\mathsf{N}$ are initialized by training on the high-resolution signals to minimize the classification loss:
\begin{equation}\label{eq5}
\underset{\mathbf{\Omega}}{\textrm{argmin}}\sum_{n=1}^{N} L\big(f_{\mathsf{N}}(\mathcal{Y}_n), \mathbf{c}_n \big),
\end{equation}
where $\mathbf{\Omega}$ denotes the parameters of $\mathsf{N}$, and $f_{\mathsf{N}}(\mathcal{Y}_n)$ denotes its prediction, given the input $\mathcal{Y}_n$. The learning loss function is denoted by $L(\cdot)$. 

Slightly related to our work is the work in \cite{tran2020performance}, which studies a surrogate performance indicator that allows fast estimation of the ranking between different model configurations in MCL. In \cite{tran2020performance}, the authors found out that the mean-squared error obtained during the initialization step in MCL can be used as a performance indicator since this quantity exhibits a high correlation with the final classification error. The work in \cite{xu2019compressed} also bears some similarity to our work in the sense that the method is also capable of learning single neural network that can classify different measurement sizes. However, the method in \cite{xu2019compressed} is different from our work since it was proposed for vector-based CL utilizing a random sensing operator, and trains the classifier based on data augmentation. Since the proposed adaptive compression MCL model relies on compressed measurements with a predefined semantic structure, the works in learning disentanglement in variational autoencoders, see for example \cite{mathieu2019disentangling}, are slightly related to our method. The main idea in learning disentanglement in generative models is to enforce certain dimensions of the feature to encode specific visual features of generated images. Different than this, our method aims to enforce an ordering structure in the compressed measurements so that multiple compressed measurements can be extracted from a single measurement.

\section{Proposed Method}\label{method}

\begin{figure*}
	\centering
	\includegraphics[width=0.8\textwidth]{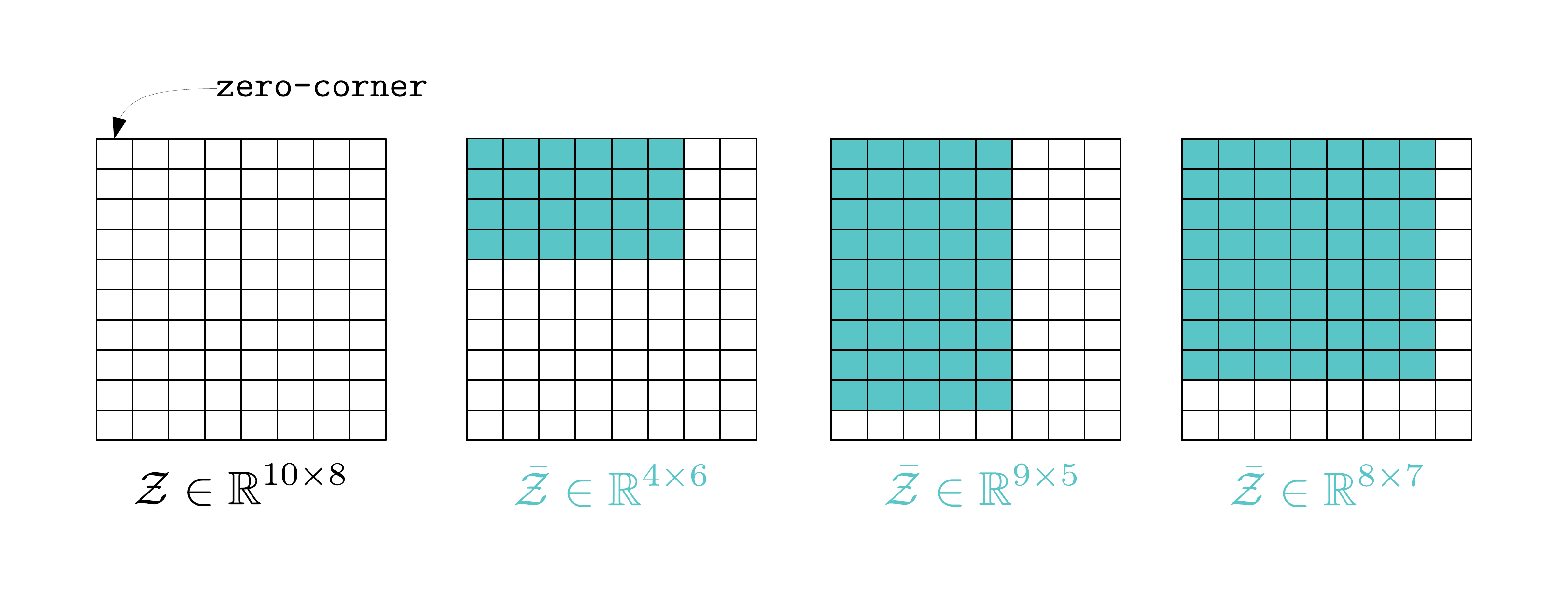}
	\caption{Illustration of compressed measurements of high compression rates $\bar{\mathcal{Z}}$ constructed from the original compressed measurement $\mathcal{Z}$.}\label{f1}
\end{figure*}

The main motivation of our work is to develop a remote Compressive Learning system that is capable of compressive signal acquisition and learning with an adaptive compression rate. The ability to adaptively adjust the compression rate, hence the degree of signal fidelity and the amount of data transmitted for each sample, can significantly enhance the information content throughput of the remote sensing and learning application. This is because in real-world scenarios, the conditions of data transmission medium, especially in wireless communication, can vary from time to time, thus, many communication protocols support adaptive transmission rates. The availability of such a feature from a communication protocol enables a network to maximize its throughput while minimizing the energy consumption. However, this feature alone cannot maximize the data content throughput, i.e., the amount of signal information transmitted in a period of time. For example, the streaming server in a video streaming service must be able to allow for an adjustable transmission rate and to send video frames with a resolution that is adaptive to the network strength in order to ensure a consistent number of frames per second, thus an uninterrupted viewing experience.

From the hardware point of view, it is infeasible to implement a MCS sensor with an adaptive compression rate through the use of multiple sets of sensing operators, each of which corresponds to a different compression rate. However, with a single set of sensing operators $\{\mathbf{\Phi}_k \, |\, k=1, \dots, K\}$ producing a compressed measurement $\mathcal{Z} \in \mathbb{R}^{M_1 \times \dots \times M_K}$, we can obtain a compressed measurement of a given size $\bar{\mathcal{Z}} \in \mathbb{R}^{m_1 \times \dots \times m_K}$ ($m_k \leq M_k, \forall k$) that corresponds to a (higher) compression rate by forming $\bar{\mathcal{Z}}$ from the elements in $\mathcal{Z}$, i.e.:
\begin{equation}\label{eq6}
\bar{\mathcal{Z}}[i_1, \dots, i_K] \in \mathcal{Z}, \quad \forall \; i_k \leq m_k \; | \; k=1, \dots, K, \\
\end{equation}
where $\bar{\mathcal{Z}}[i_1, \dots, i_K]$ denotes the element of $\bar{\mathcal{Z}}$ at position $(i_1, \dots, i_K)$. 
To construct multiple instances of $\bar{\mathcal{Z}}$, each of which carries the amount of signal information approximately proportional to its size, we can optimize the set of sensing operators $\{\mathbf{\Phi}_k \, |\, k=1, \dots, K\}$ in such a way that the resulting $\mathcal{Z}$ possesses a predefined semantic structure. 
Specifically, we aim to learn a set of sensing operators that results in $\mathcal{Z}$ such that elements in $\mathcal{Z}$ carrying the most relevant signal information for the learning task concentrate around the zero-corner of $\mathcal{Z}$, i.e., the corner at position $(1, \dots, 1)$. Furthermore, the elements are arranged according to their importance, with elements closer to position $(1, \dots, 1)$ being more relevant. 


With $\mathcal{Z}$ having the aforementioned structure, a compressed measurement $\bar{\mathcal{Z}}$ corresponding to a higher compression rate is a sub-tensor of $\mathcal{Z}$, that is:
\begin{equation}\label{eq7}
\bar{\mathcal{Z}}[i_1, \dots, i_K] = \mathcal{Z}[i_1, \dots, i_K], \:\: \forall \; i_k \leq m_k \; | \; k=1, \dots, K.
\end{equation}
The construction of $\bar{\mathcal{Z}}$ is illustrated in Figure \ref{f1}. One feature of the proposed semantic structure for $\mathcal{Z}$ is its computational efficiency. Since this structure allows the construction of $\bar{\mathcal{Z}}$ from contiguous elements of $\mathcal{Z}$, the indexing operation needed to create $\mathcal{Z}$ only requires accessing contiguous memory locations, a process that is more hardware-friendly compared to using a set of non-contiguous indices.    
\begin{figure*}
	\centering
	\includegraphics[width=0.9\textwidth]{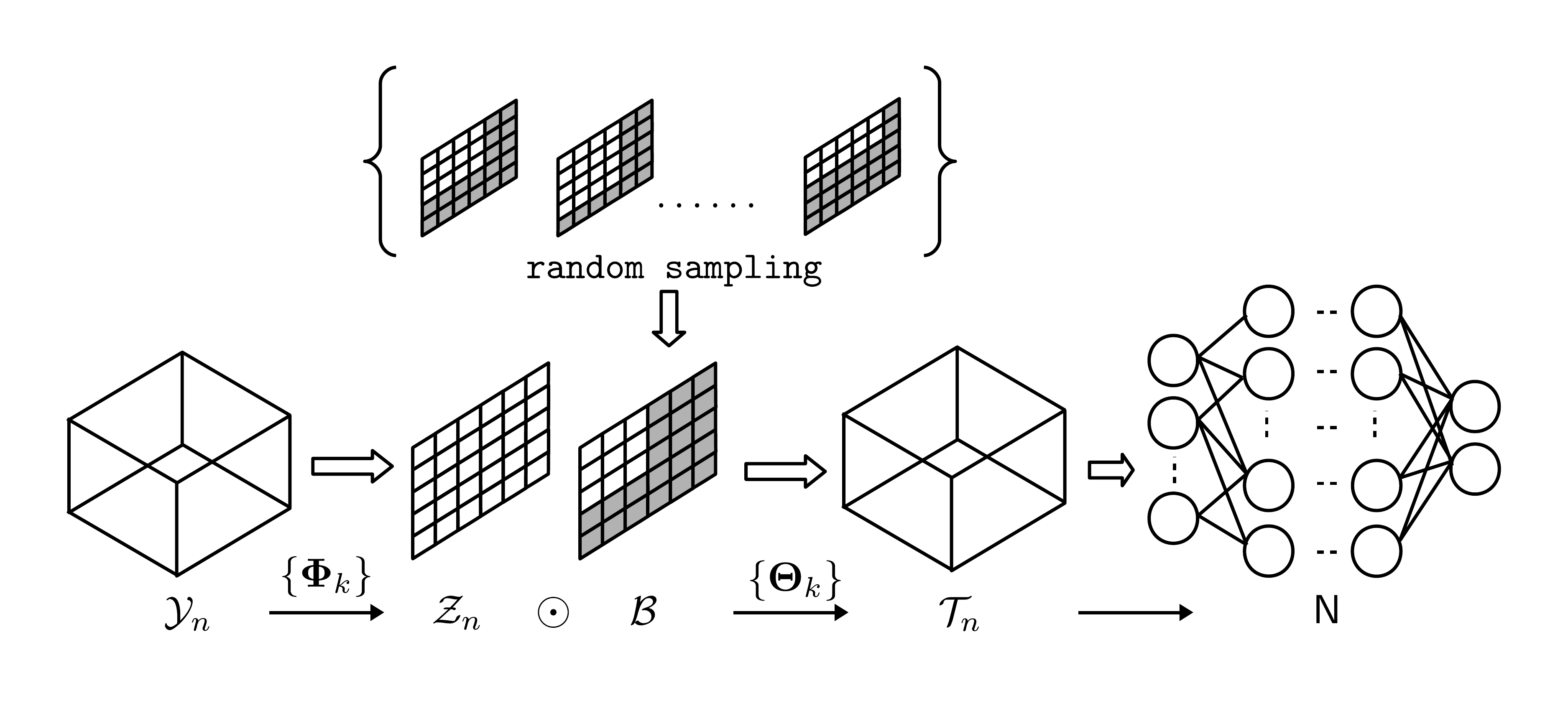}
	\caption{Illustration of the proposed training method with stochastic binary mask $\mathcal{B}$.}\label{f2}
\end{figure*}

Here we should note that $\bar{\mathcal{Z}}$ is constructed on the client side, before being transmitted to the  server. On the server side, where the FS module and task-specific neural network $\mathsf{N}$ are implemented to make predictions with incoming compressed measurements of different sizes using a single instance of the FS module and $\mathsf{N}$, the FS module must be able to handle variable-size inputs. A simple solution to this requirement is to set the input size of the FS module to the maximum size of incoming compressed measurements, i.e., the size of $\mathcal{Z}$, and the incoming compressed measurements are  appropriately  zero padded to form tensors of a fixed size.

To this end, we define the following criteria for optimizing a remote MCL system: (i) the sensing step produces $\mathcal{Z}$ that has the proposed semantic structure, and (ii) the server side utilizes a single model instance to make predictions with variable-size compressed measurements. Here we should note that the initialization step (using HOSVD) in MCL, by itself, cannot induce the proposed structures in the compressed measurements after the whole model has been optimized. This is because during stochastic optimization, all parameters are jointly updated to optimize the learning objective and there is no explicit constraint to induce such a feature. This is evidenced in our experiments in Section IV. In order to satisfy both criteria using end-to-end training with stochastic gradient descent, we propose to randomly simulate the effect of variable compression rates via the means of stochastic structural dropout. Dropout \cite{srivastava2014dropout}, which randomly zeroes out intermediate representations in a neural network during optimization, is a regularization technique. This technique can be considered as training a virtual ensemble of sub-networks within the main model, thus, it is effective at reducing overfitting. In our case, we use dropout masks having predefined structures not only to train sub-networks that correspond to different compression rates but also to enforce a semantic structure in $\mathcal{Z}$. More specifically, after performing the initialization steps of MCL described in Section \ref{related-works}, a stochastic gradient descent-based optimizer is used to train all components in MCL with the following objective: 
\begin{equation}\label{eq8}
	\underset{\{\mathbf{\Phi}_k\}, \{\mathbf{\Theta}_k\}, \mathbf{\Omega}}{\textrm{argmin}} \quad \sum_{n=1}^{N}L\Big(f_{\mathsf{N}}(f_{\mathsf{FS}}(\mathcal{Z}_n \odot \mathcal{B})), \mathbf{c}_n \Big)
\end{equation}
where $\odot$ denotes the element-wise multiplication operator. $\mathcal{B} \in \mathbb{R}^{M_1 \times \dots \times M_K}$ is a random binary matrix having the following structure:

\begin{equation}\label{eq9}
	\mathcal{B}[i_1, \dots, i_K] = \begin{dcases} 1 \quad \textrm{if }  i_k \leq m_k^{(r)} \\
		0 \quad \textrm{else}
	\end{dcases}  \quad \; \forall k=1, \dots, K
\end{equation}
where $m_k^{(r)}$ denotes an integer value randomly drawn from the set $\{M_k^{(\textrm{min})}, M_k^{(\textrm{min})} +1,\dots, M_k\}$ for all $k=1, \dots, K$, with $M_k^{(\textrm{min})}$ and $M_k$ denoting predefined minimum and maximum values for a given dimension of the compressed measurement. In practice, based on the specifications of the transmission network, we can always estimate suitable values for $M_k$ and $M_k^{(\textrm{min})}$ for all $k$.

The proposed training process with stochastic binary mask $\mathcal{B}$ is illustrated in Figure \ref{f2}. The proposed dropout strategy is a simple, yet efficient way to express our goal of inducing the aforementioned semantic structure in the optimization process: with the stochastic dropout mask, we simply instruct the stochastic optimizer to optimize all parameters of the model so that classification error is minimized for \textit{any compressed measurement size that lies within the minimum and maximum sizes.} By applying the binary mask $\mathcal{B}$ to $\mathcal{Z}_n$, we implicitly train the MCL model to perform sensing and learning with a compressed measurement of size $m_1^{(r)} \times \dots \times m_k^{(r)} \dots \times m_K^{(r)}$, which is randomly defined during stochastic optimization. To do so, the value of $\mathcal{B}$ is changed in every forward pass for every training sample during the optimization.

Finally, we should note that in order to adopt the proposed MCL with adaptive compression rate, the server side is not constrained to use a single instance of the FS module and task-specific neural network. While the client side, i.e., the sensing device, has critical limitation in terms of computational power and energy, there is no inherent limitation for the  server. In case the server has enough computational power, we can increase the performance of the entire system by running multiple FS modules and task-specific neural networks to make predictions for different compressed measurement sizes. That is, after optimizing the parameters of the entire MCL system using the loss function in Eq. (\ref{eq8}), we can fix the sensing operators and finetune the parameters of the components at the server side (FS and $\mathsf{N}$) for each compression rate. During the deployment, on the sensing device, we still implement a single set of sensing operators and compressed measurements of different sizes are constructed adaptively with sub-tensors of the output of the sensing device as described previously. On the server side, based on the size of the received compressed measurement, prediction is generated with the corresponding FS and $\mathsf{N}$ modules. In Section \ref{experiments}, we show that this approach can significantly enhance the overall performance of the system.

\section{Experiments}\label{experiments}

\begin{table}[t]
\centering
	\caption{Test Accuracy (\%) on CIFAR10}
	\label{t1}
\resizebox{0.95\linewidth}{!}{%
\begin{tabular}{c|c|c|c|}
\cline{2-4}
	\multicolumn{1}{l|}{}                       & \multirow{2}{*}{\makecell{\texttt{single}\\ \texttt{-rate}}} & \multicolumn{2}{c|}{\texttt{one-shot}} \\ \cline{3-4} 
	\multicolumn{1}{l|}{}                       &                              & \texttt{baseline}    & \makecell{\texttt{adaptive} \\ \texttt{-rate}}   \\ \hline 
\multicolumn{1}{|c|}{$4\times 6\times 2$}   & $64.92$\scalebox{0.7}{$\pm 00.89 $}                        & $19.48$\scalebox{0.7}{$\pm 02.69 $}       & $61.08$\scalebox{0.7}{$\pm 00.30 $}           \\ \hline
\multicolumn{1}{|c|}{$6\times 4\times 2$}   & $64.83$\scalebox{0.7}{$\pm 00.21 $}                        & $20.16$\scalebox{0.7}{$\pm 00.74 $}       & $61.25$\scalebox{0.7}{$\pm 00.37 $}           \\ \hline
\multicolumn{1}{|c|}{$6\times 8\times 1$}   & $62.87$\scalebox{0.7}{$\pm 00.54 $}                        & $23.27$\scalebox{0.7}{$\pm 02.78 $}       & $61.28$\scalebox{0.7}{$\pm 00.13 $}           \\ \hline
\multicolumn{1}{|c|}{$8\times 6\times 1$}   & $62.40$\scalebox{0.7}{$\pm 00.23 $}                        & $23.25$\scalebox{0.7}{$\pm 00.56 $}       & $61.41$\scalebox{0.7}{$\pm 00.25 $}           \\ \hline \hline
\multicolumn{1}{|c|}{$4\times 7\times 2$}   & $66.28$\scalebox{0.7}{$\pm 00.16 $}                        & $20.34$\scalebox{0.7}{$\pm 03.00 $}       & $62.92$\scalebox{0.7}{$\pm 00.16 $}           \\ \hline
\multicolumn{1}{|c|}{$7\times 4\times 2$}   & $66.85$\scalebox{0.7}{$\pm 00.55 $}                        & $21.38$\scalebox{0.7}{$\pm 00.14 $}       & $63.36$\scalebox{0.7}{$\pm 00.21 $}           \\ \hline
\multicolumn{1}{|c|}{$7\times 8\times 1$}   & $65.02$\scalebox{0.7}{$\pm 00.17 $}                        & $26.75$\scalebox{0.7}{$\pm 02.89 $}       & $63.46$\scalebox{0.7}{$\pm 00.52 $}           \\ \hline
\multicolumn{1}{|c|}{$8\times 7\times 1$}   & $64.87$\scalebox{0.7}{$\pm 00.13 $}                        & $26.54$\scalebox{0.7}{$\pm 01.90 $}       & $63.86$\scalebox{0.7}{$\pm 00.65 $}           \\ \hline \hline 
\multicolumn{1}{|c|}{$6\times 6\times 2$}   & $69.87$\scalebox{0.7}{$\pm 00.20 $}                        & $26.42$\scalebox{0.7}{$\pm 04.00 $}      & $68.98$\scalebox{0.7}{$\pm 00.15 $}           \\ \hline
\multicolumn{1}{|c|}{$8\times 9\times 1$}   & $68.17$\scalebox{0.7}{$\pm 00.27 $}                        & $33.93$\scalebox{0.7}{$\pm 02.57 $}       & $67.28$\scalebox{0.7}{$\pm 00.34 $}           \\ \hline
\multicolumn{1}{|c|}{$9\times 8\times 1$}   & $67.69$\scalebox{0.7}{$\pm 00.41 $}                        & $33.62$\scalebox{0.7}{$\pm 02.89 $}       & $67.52$\scalebox{0.7}{$\pm 00.14 $}           \\ \hline
\multicolumn{1}{|c|}{$12\times 6\times 1$}  & $67.29$\scalebox{0.7}{$\pm 00.25 $}                        & $25.91$\scalebox{0.7}{$\pm 00.63 $}       & $65.72$\scalebox{0.7}{$\pm 00.23 $}           \\ \hline
\multicolumn{1}{|c|}{$9\times 4\times 2$}   & $69.20$\scalebox{0.7}{$\pm 00.33 $}                        & $22.88$\scalebox{0.7}{$\pm 02.03 $}       & $66.16$\scalebox{0.7}{$\pm 00.25 $}           \\ \hline \hline
\multicolumn{1}{|c|}{$10\times 10\times 1$} & $71.45$\scalebox{0.7}{$\pm 00.20 $}                        & $47.19$\scalebox{0.7}{$\pm 01.61 $}       & $71.42$\scalebox{0.7}{$\pm 00.28 $}           \\ \hline
\multicolumn{1}{|c|}{$10\times 5\times 2$}  & $73.66$\scalebox{0.7}{$\pm 00.77 $}                        & $29.83$\scalebox{0.7}{$\pm 00.91 $}       & $72.57$\scalebox{0.7}{$\pm 00.48$}           \\ \hline \hline
\multicolumn{1}{|c|}{average}               & $67.02$                        & $26.73$       & $65.22$           \\ \hline \hline
\multicolumn{1}{|c|}{$\sum$epochs}               & $3150$                        & $210$       & $210$           \\ \hline
\end{tabular}%
}
\end{table}

This section provides the empirical analysis conducted to benchmark our training scheme for MCL models. Information about the datasets and experimental protocol are presented first, followed by the experimental results and discussion. 

\subsection{Datasets and Experiment Protocol}

Our experiments were conducted using publicly available image datasets describing object classification and face recognition tasks. Object and face recognition are necessary features in smart buildings and surveillance systems. As we mentioned before, in order to train end-to-end compressive learning models, we need labeled data that has been collected by standard sensors. For this reason, we used CIFAR \cite{krizhevsky2009learning} and CelebA datasets \cite{liu2015faceattributes} in our experiments, both of which are publicly available. A brief description of the datasets and our data split are provided below: 

\begin{itemize}
	\item CIFAR dataset \cite{krizhevsky2009learning} is an RGB image dataset, which contains thumbnail-size images of resolution $32\times 32$ pixels. The dataset is formed by $60$K images that are divided into the training set of $50$K images and the test set of $10$K. The dataset contains two label sets, each of which has $10$ and $100$ classes, respectively. We refer to the two versions as CIFAR10 and CIFAR100. In order to perform proper validation, we randomly selected $5$K images from the training set to form the validation set. Images in different classes are uniformly distributed in both CIFAR10 and CIFAR100. 

	\item CelebA \cite{liu2015faceattributes} is a large-scale face attributes dataset with more than $200\mathrm{K}$ images of about $10K$ different people at varying resolutions. For this dataset, we followed the same experimental protocol as in \cite{tran2019multilinear} and used a subset of $100$ people to train and evaluate the performance of the methods. The training, validation and test sets contain $7063$, $2373$ and $2400$ images, respectively. All images were resized to fixed resolution images of $32\times 32$ pixels. 	

\end{itemize}

\begin{table}[t]
\centering
	\caption{Test Accuracy (\%) on CIFAR100}
	\label{t2}
\resizebox{0.95\linewidth}{!}{%
\begin{tabular}{c|c|c|c|}
\cline{2-4}
	\multicolumn{1}{l|}{}                       & \multirow{2}{*}{\makecell{\texttt{single}\\ \texttt{-rate}}} & \multicolumn{2}{c|}{\texttt{one-shot}} \\ \cline{3-4} 
	\multicolumn{1}{l|}{}                       &                              & \texttt{baseline}    & \makecell{\texttt{adaptive} \\ \texttt{-rate}}   \\ \hline 
\multicolumn{1}{|c|}{$4\times 6\times 2$}   & $36.13$\scalebox{0.7}{$\pm 00.88 $}                        & $03.33$\scalebox{0.7}{$\pm 01.44 $}       & $33.05$\scalebox{0.7}{$\pm 00.71 $}           \\ \hline
\multicolumn{1}{|c|}{$6\times 4\times 2$}   & $36.69$\scalebox{0.7}{$\pm 00.18 $}                        & $03.64$\scalebox{0.7}{$\pm 00.91 $}       & $35.77$\scalebox{0.7}{$\pm 00.36 $}           \\ \hline
\multicolumn{1}{|c|}{$6\times 8\times 1$}   & $30.84$\scalebox{0.7}{$\pm 00.26 $}                        & $04.96$\scalebox{0.7}{$\pm 00.87 $}       & $30.53$\scalebox{0.7}{$\pm 00.92 $}           \\ \hline
\multicolumn{1}{|c|}{$8\times 6\times 1$}   & $30.93$\scalebox{0.7}{$\pm 00.33 $}                        & $04.37$\scalebox{0.7}{$\pm 01.39 $}       & $30.92$\scalebox{0.7}{$\pm 00.08 $}           \\ \hline \hline
\multicolumn{1}{|c|}{$4\times 7\times 2$}   & $38.26$\scalebox{0.7}{$\pm 00.91 $}                        & $04.10$\scalebox{0.7}{$\pm 01.47 $}       & $35.44$\scalebox{0.7}{$\pm 01.26 $}           \\ \hline
\multicolumn{1}{|c|}{$7\times 4\times 2$}   & $38.19$\scalebox{0.7}{$\pm 00.27 $}                        & $04.36$\scalebox{0.7}{$\pm 02.14 $}       & $37.66$\scalebox{0.7}{$\pm 00.39 $}           \\ \hline
\multicolumn{1}{|c|}{$7\times 8\times 1$}   & $32.94$\scalebox{0.7}{$\pm 00.29 $}                        & $06.22$\scalebox{0.7}{$\pm 00.92 $}       & $32.78$\scalebox{0.7}{$\pm 00.16 $}           \\ \hline
\multicolumn{1}{|c|}{$8\times 7\times 1$}   & $32.65$\scalebox{0.7}{$\pm 00.15 $}                        & $05.93$\scalebox{0.7}{$\pm 01.64 $}       & $33.15$\scalebox{0.7}{$\pm 00.03 $}           \\ \hline \hline
\multicolumn{1}{|c|}{$6\times 6\times 2$}   & $39.67$\scalebox{0.7}{$\pm 01.42 $}                        & $05.19$\scalebox{0.7}{$\pm 02.25 $}       & $41.08$\scalebox{0.7}{$\pm 00.20 $}           \\ \hline
\multicolumn{1}{|c|}{$8\times 9\times 1$}   & $35.63$\scalebox{0.7}{$\pm 00.17 $}                        & $09.45$\scalebox{0.7}{$\pm 00.86 $}       & $36.52$\scalebox{0.7}{$\pm 00.13 $}           \\ \hline
\multicolumn{1}{|c|}{$9\times 8\times 1$}   & $35.63$\scalebox{0.7}{$\pm 01.36 $}                        & $09.22$\scalebox{0.7}{$\pm 00.87 $}       & $36.04$\scalebox{0.7}{$\pm 00.10 $}           \\ \hline
\multicolumn{1}{|c|}{$12\times 6\times 1$}  & $34.82$\scalebox{0.7}{$\pm 00.30 $}                        & $07.36$\scalebox{0.7}{$\pm 01.74 $}       & $34.59$\scalebox{0.7}{$\pm 00.19 $}           \\ \hline
\multicolumn{1}{|c|}{$9\times 4\times 2$}   & $40.73$\scalebox{0.7}{$\pm 01.02 $}                        & $05.40$\scalebox{0.7}{$\pm 02.12 $}       & $40.43$\scalebox{0.7}{$\pm 00.15 $}           \\ \hline \hline
\multicolumn{1}{|c|}{$10\times 10\times 1$} & $39.82$\scalebox{0.7}{$\pm 00.89 $}                        & $16.54$\scalebox{0.7}{$\pm 01.37 $}       & $40.34$\scalebox{0.7}{$\pm 00.14 $}           \\ \hline
\multicolumn{1}{|c|}{$10\times 5\times 2$}  & $44.82$\scalebox{0.7}{$\pm 00.53 $}                        & $07.05$\scalebox{0.7}{$\pm 00.85 $}       & $44.86$\scalebox{0.7}{$\pm 00.24$}           \\ \hline \hline
\multicolumn{1}{|c|}{average}               & $36.52$                        & $06.47$       & $36.21$           \\ \hline \hline
\multicolumn{1}{|c|}{$\sum$epochs}               & $3150$                        & $210$       & $210$           \\ \hline
\end{tabular}%
}
\end{table}

In our experiments, we adopted the same task-specific neural network architecture that was used in the experiments of \cite{tran2019multilinear}, namely the AllCNN architecture proposed by \cite{springenberg2014striving}. AllCNN is a feed-forward architecture that contains only convolution layers, without any residual connection. For the details of AllCNN network, the reader is referred  to \cite{tran2019multilinear}. 

For stochastic gradient descent optimization, we used ADAM optimizer \cite{kingma2014adam} with $\beta_1 = 0.9$ and $\beta_2 = 0.999$. All models were trained for $210$ epochs with an initial learning rate of $0.001$, which is reduced by a factor of $10$ at epoch $51$ and $190$, respectively. Weight decay of magnitude $0.00005$ was used for regularization. The input pixel values were scaled into the range $[0, 1]$, and we performed a simple data augmentation technique during training by random horizontal flipping and random shifting by $4$ pixels in both horizontal and vertical axes. We tested many configurations corresponding to a variety of compression rates. The experiments were repeated five times and  the means and standard deviations of the accuracy measured on the test set are reported. 

\subsection{Results}

Since CIFAR10, CIFAR100 and CelebA datasets were set to the same resolution, the size of $\mathcal{Y}_n$ is $32\times 32\times 3$ in all experiments. In the following, we will denote the results produced by the original training method proposed in \cite{tran2019multilinear} as \texttt{single-rate}, and our one-shot training method as \texttt{adaptive-rate}. 

For the \texttt{single-rate} training we trained multiple models, each for one of the following compressed measurement sizes: $4\times 6\times 2$, $6\times 4\times 2$, $6\times 8\times 1$, $8\times 6\times 1$; $4\times 7\times 2$, $7\times 4\times 2$, $7\times 8\times 1$, $8\times 7\times 1$; $6\times 6\times 2$, $8\times 9\times 1$, $9\times 8\times 1$, $12\times 6\times 1$, $9\times 4\times 2$; $10\times 10\times 1$, $10\times 5\times 2$. 

\begin{table}[t]
\centering
	\caption{Test Accuracy (\%) on CelebA}
	\label{t3}
\resizebox{0.95\linewidth}{!}{%
\begin{tabular}{c|c|c|c|}
\cline{2-4}
	\multicolumn{1}{l|}{}                       & \multirow{2}{*}{\makecell{\texttt{single}\\ \texttt{-rate}}} & \multicolumn{2}{c|}{\texttt{one-shot}} \\ \cline{3-4} 
	\multicolumn{1}{l|}{}                       &                              & \texttt{baseline}    & \makecell{\texttt{adaptive} \\ \texttt{-rate}}   \\ \hline 
\multicolumn{1}{|c|}{$4\times 6\times 2$}   & $49.12$\scalebox{0.7}{$\pm 00.17$}                        & $06.42$\scalebox{0.7}{$\pm 00.17$}       & $45.60$\scalebox{0.7}{$\pm 00.31$}           \\ \hline
\multicolumn{1}{|c|}{$6\times 4\times 2$}   & $53.38$\scalebox{0.7}{$\pm 00.13$}                        & $08.02$\scalebox{0.7}{$\pm 00.52$}       & $51.37$\scalebox{0.7}{$\pm 00.17$}           \\ \hline
\multicolumn{1}{|c|}{$6\times 8\times 1$}   & $61.00$\scalebox{0.7}{$\pm 00.17$}                        & $09.65$\scalebox{0.7}{$\pm 02.66$}       & $57.60$\scalebox{0.7}{$\pm 00.48$}           \\ \hline
\multicolumn{1}{|c|}{$8\times 6\times 1$}   & $63.38$\scalebox{0.7}{$\pm 00.88$}                        & $16.60$\scalebox{0.7}{$\pm 00.52$}       & $62.12$\scalebox{0.7}{$\pm 00.54$}           \\ \hline \hline
\multicolumn{1}{|c|}{$4\times 7\times 2$}   & $51.79$\scalebox{0.7}{$\pm 00.24$}                        & $07.06$\scalebox{0.7}{$\pm 00.12$}       & $49.33$\scalebox{0.7}{$\pm 00.74$}           \\ \hline
\multicolumn{1}{|c|}{$7\times 4\times 2$}   & $60.17$\scalebox{0.7}{$\pm 00.12$}                        & $11.42$\scalebox{0.7}{$\pm 00.13$}       & $56.56$\scalebox{0.7}{$\pm 00.35$}           \\ \hline
\multicolumn{1}{|c|}{$7\times 8\times 1$}   & $68.08$\scalebox{0.7}{$\pm 00.37$}                        & $16.31$\scalebox{0.7}{$\pm 01.31$}       & $64.21$\scalebox{0.7}{$\pm 00.25$}           \\ \hline
\multicolumn{1}{|c|}{$8\times 7\times 1$}   & $68.75$\scalebox{0.7}{$\pm 00.82$}                        & $17.58$\scalebox{0.7}{$\pm 00.12$}       & $64.77$\scalebox{0.7}{$\pm 00.56$}           \\ \hline \hline
\multicolumn{1}{|c|}{$6\times 6\times 2$}   & $60.40$\scalebox{0.7}{$\pm 01.54$}                        & $13.17$\scalebox{0.7}{$\pm 01.37$}       & $58.00$\scalebox{0.7}{$\pm 00.23$}           \\ \hline
\multicolumn{1}{|c|}{$8\times 9\times 1$}   & $73.06$\scalebox{0.7}{$\pm 00.19$}                        & $23.50$\scalebox{0.7}{$\pm 00.58$}       & $69.44$\scalebox{0.7}{$\pm 00.17$}           \\ \hline
\multicolumn{1}{|c|}{$9\times 8\times 1$}   & $74.40$\scalebox{0.7}{$\pm 00.52$}                        & $28.35$\scalebox{0.7}{$\pm 00.77$}       & $71.31$\scalebox{0.7}{$\pm 00.39$}           \\ \hline
\multicolumn{1}{|c|}{$12\times 6\times 1$}  & $74.48$\scalebox{0.7}{$\pm 00.23$}                        & $30.92$\scalebox{0.7}{$\pm 01.46$}       & $72.23$\scalebox{0.7}{$\pm 00.29$}           \\ \hline
\multicolumn{1}{|c|}{$9\times 4\times 2$}   & $66.40$\scalebox{0.7}{$\pm 00.19$}                        & $17.02$\scalebox{0.7}{$\pm 00.15$}       & $64.44$\scalebox{0.7}{$\pm 00.35$}           \\ \hline \hline
\multicolumn{1}{|c|}{$10\times 10\times 1$} & $80.00$\scalebox{0.7}{$\pm 00.21$}                        & $44.71$\scalebox{0.7}{$\pm 01.88$}       & $76.75$\scalebox{0.7}{$\pm 00.29$}           \\ \hline
\multicolumn{1}{|c|}{$10\times 5\times 2$}  & $72.92$\scalebox{0.7}{$\pm 00.25$}                        & $22.73$\scalebox{0.7}{$\pm 02.02$}       & $70.00$\scalebox{0.7}{$\pm 00.41$}           \\ \hline \hline
\multicolumn{1}{|c|}{average}               & $65.16$                        & $18.23$       & $62.25$           \\ \hline \hline
\multicolumn{1}{|c|}{$\sum$epochs}               & $3150$                        & $210$       & $210$           \\ \hline
\end{tabular}%
}
\end{table}

For our \texttt{adaptive-rate} training method, we trained only one model with the maximum and minimum dimensions of the compressed measurements set to $15\times 15\times 2$ and $4\times 4\times 1$. That is, the size of $\mathcal{Z}_n$ in Eq. (\ref{eq8}) is ${15\times 15\times 2}$, and $M_1^{(min)} = M_2^{(min)} = 4$ and $M_3^{(min)} = 1$ when sampling $\mathcal{B}$ in Eq. (\ref{eq9}). After training, we simply evaluated this model with different compressed measurement sizes that were used to train the \texttt{single-rate} models.

In addition, to demonstrate the effectiveness of the stochastic mask $\mathcal{B}$, we also trained a MCL model with the compressed measurement of size $15\times 15\times 2$, using the original training method in \cite{tran2019multilinear}. This model, denoted as \texttt{baseline}, is then used to evaluate the target set of compressed measurements mentioned above, using the same evaluation procedure as in \texttt{adaptive-rate} models. Thus, \texttt{baseline} model and our (\texttt{adaptive-rate}) model have the same setup that represents an one-shot training setting in which one model is trained and used for multiple compressed measurement sizes.    
The results for CIFAR10, CIFAR100 and CelebA datasets are shown in Tables \ref{t1}, \ref{t2} and \ref{t3}, respectively. The results are grouped according to the compression rate. The average accuracy of each method and the total number of epochs used to train the model(s) used for each dataset are also provided in these tables.  

\begin{table}[t]
\centering
	\caption{Finetuning Performance of \lowercase{\texttt{adaptive-rate*}} on CIFAR10}
	\label{t4}
\resizebox{0.95\linewidth}{!}{%
\begin{tabular}{c|c|c|c|}
\cline{2-4}
\multicolumn{1}{l|}{}                       & \makecell{\texttt{single}\\ \texttt{-rate}} & \makecell{\texttt{adaptive}\\ \texttt{-rate}} &  \makecell{\texttt{adaptive}\\ \texttt{-rate*}}\\ \hline
\multicolumn{1}{|c|}{$4\times 6\times 2$}   & $64.92$\scalebox{0.7}{$\pm 00.89$}   & $61.08$\scalebox{0.7}{$\pm 00.30$}     & $65.22$\scalebox{0.7}{$\pm 00.20$}     \\ \hline
\multicolumn{1}{|c|}{$6\times 4\times 2$}   & $64.83$\scalebox{0.7}{$\pm 00.21$}   & $61.25$\scalebox{0.7}{$\pm 00.37$}     & $65.48$\scalebox{0.7}{$\pm 00.89$}     \\ \hline
\multicolumn{1}{|c|}{$6\times 8\times 1$}   & $62.87$\scalebox{0.7}{$\pm 00.54$}   & $61.28$\scalebox{0.7}{$\pm 00.13$}     & $62.97$\scalebox{0.7}{$\pm 00.44$}     \\ \hline
\multicolumn{1}{|c|}{$8\times 6\times 1$}   & $62.40$\scalebox{0.7}{$\pm 00.23$}   & $61.41$\scalebox{0.7}{$\pm 00.25$}     & $62.73$\scalebox{0.7}{$\pm 00.29$}     \\ \hline \hline
\multicolumn{1}{|c|}{$4\times 7\times 2$}   & $66.28$\scalebox{0.7}{$\pm 00.16$}   & $62.92$\scalebox{0.7}{$\pm 00.16$}     & $67.11$\scalebox{0.7}{$\pm 00.78$}     \\ \hline
\multicolumn{1}{|c|}{$7\times 4\times 2$}   & $66.85$\scalebox{0.7}{$\pm 00.55$}   & $63.36$\scalebox{0.7}{$\pm 00.21$}     & $67.74$\scalebox{0.7}{$\pm 00.41$}     \\ \hline
\multicolumn{1}{|c|}{$7\times 8\times 1$}   & $65.02$\scalebox{0.7}{$\pm 00.17$}   & $63.46$\scalebox{0.7}{$\pm 00.52$}     & $65.22$\scalebox{0.7}{$\pm 00.45$}     \\ \hline
\multicolumn{1}{|c|}{$8\times 7\times 1$}   & $64.87$\scalebox{0.7}{$\pm 00.13$}   & $63.86$\scalebox{0.7}{$\pm 00.65$}     & $65.09$\scalebox{0.7}{$\pm 00.79$}     \\ \hline \hline
\multicolumn{1}{|c|}{$6\times 6\times 2$}   & $69.87$\scalebox{0.7}{$\pm 00.20$}   & $68.98$\scalebox{0.7}{$\pm 00.15$}     & $70.62$\scalebox{0.7}{$\pm 00.53$}     \\ \hline
\multicolumn{1}{|c|}{$8\times 9\times 1$}   & $68.17$\scalebox{0.7}{$\pm 00.27$}   & $67.28$\scalebox{0.7}{$\pm 00.34$}     & $68.61$\scalebox{0.7}{$\pm 00.41$}     \\ \hline
\multicolumn{1}{|c|}{$9\times 8\times 1$}   & $67.69$\scalebox{0.7}{$\pm 00.41$}   & $67.52$\scalebox{0.7}{$\pm 00.14$}     & $68.45$\scalebox{0.7}{$\pm 00.82$}     \\ \hline
\multicolumn{1}{|c|}{$12\times 6\times 1$}  & $67.29$\scalebox{0.7}{$\pm 00.25$}   & $65.72$\scalebox{0.7}{$\pm 00.23$}     & $67.76$\scalebox{0.7}{$\pm 00.43$}     \\ \hline
\multicolumn{1}{|c|}{$9\times 4\times 2$}   & $69.20$\scalebox{0.7}{$\pm 00.33$}   & $66.16$\scalebox{0.7}{$\pm 00.25$}     & $70.35$\scalebox{0.7}{$\pm 00.45$}     \\ \hline \hline
\multicolumn{1}{|c|}{$10\times 10\times 1$} & $71.45$\scalebox{0.7}{$\pm 00.20$}   & $71.42$\scalebox{0.7}{$\pm 00.28$}     & $72.58$\scalebox{0.7}{$\pm 00.91$}     \\ \hline
\multicolumn{1}{|c|}{$10\times 5\times 2$}  & $73.66$\scalebox{0.7}{$\pm 00.77$}   & $72.57$\scalebox{0.7}{$\pm 00.48$}     & $74.63$\scalebox{0.7}{$\pm 00.75$}     \\ \hline \hline
	\multicolumn{1}{|c|}{average}               & $67.02$                         & $65.22$   & $67.64$        \\ \hline \hline
\multicolumn{1}{|c|}{$\sum$epochs}               & $3150$                        & $210$       & $660$           \\ \hline
\end{tabular}%
}
\end{table}

As can be seen in Tables \ref{t1}, \ref{t2} and \ref{t3}, the proposed method (\texttt{adaptive-rate}) clearly outperforms the \texttt{baseline} method. In fact, we can see that without any modification to the original \texttt{baseline} training algorithm, the model trained with a large compressed measurement size ($15\times 15\times 2$ in our case) cannot be used for other configurations with higher compression rates (i.e. smaller measurement sizes). The large variations between different runs of the \texttt{baseline} indicate that there is no semantic structure existing in the sub-tensors of the default compressed measurement, i.e., the compressed measurement of the maximum size. On the other hand, the results from \texttt{adaptive-rate} are more consistent between different experiment runs. This means that from the default compressed measurement of size $15\times 15\times 2$, we can directly generate compressed measurements of smaller sizes which lead to consistent performance, indicating the existence of a semantic structure between elements in the compressed measurement trained by the proposed method.

Regarding the comparison with the conventional approach of \texttt{single-rate} using multiple models, each specializing on a compression rate, we can see that one model trained using the proposed \texttt{adaptive-rate} approach achieved very competitive performance. Specifically, the performance of the model trained using the \texttt{adaptive-rate} approach achieved performance close to that of the models trained using the \texttt{single-rate} approach on CIFAR100 dataset, while its performance is slightly lower on CIFAR10 and CelebA datasets, with an average of $1.8\%$ and $2.91\%$ performance degradation, respectively. However, the slight performance loss in \texttt{adaptive-rate} is compensated with significant computational and memory gains. For the \texttt{single-rate} approach, $15$ different compressed measurements need to be deployed, which are trained for a total of $3150$ epochs of gradient updates. On the other hand, with \texttt{adaptive-rate} training, we needed to train only one model, which can be used for all $15$ different configurations corresponding to multiple compression rates. 

Comparing the inference complexity between the \texttt{adaptive-rate} and a \texttt{single-rate} model, the \texttt{adaptive-rate} model requires slightly more floating-point operations. However, the difference resulted from the difference in the MCS and FS components, is very minor because the majority of computation happens in the task-specific neural network. For example, the AllCNN task-specific neural network requires about $161$ millions floating-point operations while the MCS and FS components require about $120K$ floating-point operations, which account for about $0.07\%$ of the amount of computation required by the task-specific neural network in our experiments. In fact, the theoretical difference is so minor that we observe no actual differences in the inference run-time. For the same reason, the computational complexity required to train an \texttt{adaptive-rate} model for one epoch is very similar to that of a \texttt{single-rate} model. Thus, we use the number of training epochs to reflect the computational complexity induced by each model. Even though the total training time can better reflect the training complexity, this measure is not a reliable estimate of the training complexity in our case since different experiments were conducted on different workstations with different GPU models.

\begin{table}[t]
\centering
	\caption{Finetuning Performance of \lowercase{\texttt{adaptive-rate*}} on CIFAR100}
	\label{t5}
\resizebox{0.95\linewidth}{!}{%
\begin{tabular}{c|c|c|c|}
\cline{2-4}
\multicolumn{1}{l|}{}                       & \makecell{\texttt{single}\\ \texttt{-rate}} & \makecell{\texttt{adaptive}\\ \texttt{-rate}} &  \makecell{\texttt{adaptive}\\ \texttt{-rate*}}\\ \hline 
\multicolumn{1}{|c|}{$4\times 6\times 2$}   & $36.13$\scalebox{0.7}{$\pm 00.88$}   & $33.05$\scalebox{0.7}{$\pm 00.71$}     & $37.27$\scalebox{0.7}{$\pm 01.03$}     \\ \hline
\multicolumn{1}{|c|}{$6\times 4\times 2$}   & $36.69$\scalebox{0.7}{$\pm 00.18$}   & $35.77$\scalebox{0.7}{$\pm 00.36$}     & $38.25$\scalebox{0.7}{$\pm 00.95$}     \\ \hline
\multicolumn{1}{|c|}{$6\times 8\times 1$}   & $30.84$\scalebox{0.7}{$\pm 00.26$}   & $30.53$\scalebox{0.7}{$\pm 00.92$}     & $31.97$\scalebox{0.7}{$\pm 00.33$}     \\ \hline
\multicolumn{1}{|c|}{$8\times 6\times 1$}   & $30.93$\scalebox{0.7}{$\pm 00.33$}   & $30.92$\scalebox{0.7}{$\pm 00.08$}     & $32.10$\scalebox{0.7}{$\pm 01.25$}     \\ \hline \hline
\multicolumn{1}{|c|}{$4\times 7\times 2$}   & $38.26$\scalebox{0.7}{$\pm 00.91$}   & $35.44$\scalebox{0.7}{$\pm 01.26$}     & $39.70$\scalebox{0.7}{$\pm 00.76$}     \\ \hline
\multicolumn{1}{|c|}{$7\times 4\times 2$}   & $38.19$\scalebox{0.7}{$\pm 00.27$}   & $37.66$\scalebox{0.7}{$\pm 00.39$}     & $40.30$\scalebox{0.7}{$\pm 00.23$}     \\ \hline
\multicolumn{1}{|c|}{$7\times 8\times 1$}   & $32.94$\scalebox{0.7}{$\pm 00.29$}   & $32.78$\scalebox{0.7}{$\pm 00.16$}     & $33.88$\scalebox{0.7}{$\pm 00.43$}     \\ \hline
\multicolumn{1}{|c|}{$8\times 7\times 1$}   & $32.65$\scalebox{0.7}{$\pm 00.15$}   & $33.15$\scalebox{0.7}{$\pm 00.03$}     & $34.04$\scalebox{0.7}{$\pm 01.09$}     \\ \hline \hline
\multicolumn{1}{|c|}{$6\times 6\times 2$}   & $39.67$\scalebox{0.7}{$\pm 01.42$}   & $41.08$\scalebox{0.7}{$\pm 00.20$}     & $42.71$\scalebox{0.7}{$\pm 00.74$}     \\ \hline
\multicolumn{1}{|c|}{$8\times 9\times 1$}   & $35.63$\scalebox{0.7}{$\pm 00.17$}   & $36.52$\scalebox{0.7}{$\pm 00.13$}     & $36.97$\scalebox{0.7}{$\pm 00.29$}     \\ \hline
\multicolumn{1}{|c|}{$9\times 8\times 1$}   & $35.63$\scalebox{0.7}{$\pm 01.36$}   & $36.04$\scalebox{0.7}{$\pm 00.10$}     & $36.87$\scalebox{0.7}{$\pm 00.17$}     \\ \hline
\multicolumn{1}{|c|}{$12\times 6\times 1$}  & $34.82$\scalebox{0.7}{$\pm 00.30$}   & $34.59$\scalebox{0.7}{$\pm 00.19$}     & $36.14$\scalebox{0.7}{$\pm 00.62$}     \\ \hline
\multicolumn{1}{|c|}{$9\times 4\times 2$}   & $40.73$\scalebox{0.7}{$\pm 01.02$}   & $40.43$\scalebox{0.7}{$\pm 00.15$}     & $42.94$\scalebox{0.7}{$\pm 00.77$}     \\ \hline \hline
\multicolumn{1}{|c|}{$10\times 10\times 1$} & $39.82$\scalebox{0.7}{$\pm 00.89$}   & $40.34$\scalebox{0.7}{$\pm 00.14$}     & $40.88$\scalebox{0.7}{$\pm 00.31$}     \\ \hline
\multicolumn{1}{|c|}{$10\times 5\times 2$}  & $44.82$\scalebox{0.7}{$\pm 00.53$}   & $44.86$\scalebox{0.7}{$\pm 00.24$}     & $46.20$\scalebox{0.7}{$\pm 00.13$}     \\ \hline \hline
	\multicolumn{1}{|c|}{average}               & $36.52$                          & $36.21$  & $38.01$           \\ \hline \hline
\multicolumn{1}{|c|}{$\sum$epochs}               & $3150$                        & $210$       & $660$           \\ \hline
\end{tabular}%
}
\end{table}

Another benefit of using the proposed training process is that we can also use \texttt{adaptive-rate} training technique to quickly identify the best configurations for a given compression rate. For example, in CelebA dataset, even though $8\times 6\times 1$ and $6\times 8\times 1$ correspond to the same compression rate with $4\times 6\times 2$ and $6\times 4\times 2$ when each configuration is optimized separately, the first two configurations lead to a much better accuracy. We can also observe this phenomenon when using \texttt{adaptive-rate} training, while requiring only $25\%$ of the computations compared to the former case. On CIFAR100, when using the same compression rate, the ranking is reversed; i.e., using $4\times 6\times 2$ and $6\times 4\times 2$ yields a much better accuracy compared to $8\times 6\times 1$ and $6\times 8\times 1$, which can also be seen from the results of \texttt{adaptive-rate}.        

Up until now, we have shown that with less than $3\%$ of accuracy drop, we can efficiently train and deploy a single model that can be used for inference with an adaptive compression rate. However, as we described at the end of Section \ref{method}, the server side is not constrained to run a single instance of FS module and task-specific neural network. That is, after optimizing a single model instance using \texttt{adaptive-rate} method, we can fix the sensing operators and finetune the FS module and task-specific neural network for each compressed measurement size. To demonstrate this, we performed finetuning for $30$ epochs for each compressed measurement size. The results, denoted as \texttt{adaptive-rate*}, are shown in Tables \ref{t4}, \ref{t5}, and \ref{t6}. It can be clearly seen that by allowing separate model instances on the server side with \texttt{adaptive-rate*}, we obtain noticeable improvements in the overall performance, which are on-par with the performance obtained when using the \texttt{single-rate} training approach on CIFAR10 and CelebA datasets, and clearly better on CIFAR100 dataset. Although the training complexity (total number of epochs) using \texttt{adaptive-rate*} is higher than using a single model instance, it is still far below the one of \texttt{single-rate}.

\begin{table}[t]
\centering
	\caption{Finetuning Performance of \lowercase{\texttt{adaptive-rate*}} on CelebA}
	\label{t6}
\resizebox{0.95\linewidth}{!}{%
\begin{tabular}{c|c|c|c|}
\cline{2-4}
\multicolumn{1}{l|}{}                       & \makecell{\texttt{single}\\ \texttt{-rate}} & \makecell{\texttt{adaptive}\\ \texttt{-rate}} &  \makecell{\texttt{adaptive}\\ \texttt{-rate*}}\\ \hline 
\multicolumn{1}{|c|}{$4\times 6\times 2$}   & $49.12$\scalebox{0.7}{$\pm 00.17$}   & $45.60$\scalebox{0.7}{$\pm 00.31$}     & $49.92$\scalebox{0.7}{$\pm 00.42$}     \\ \hline
\multicolumn{1}{|c|}{$6\times 4\times 2$}   & $53.38$\scalebox{0.7}{$\pm 00.13$}   & $51.37$\scalebox{0.7}{$\pm 00.17$}     & $53.85$\scalebox{0.7}{$\pm 00.15$}     \\ \hline
\multicolumn{1}{|c|}{$6\times 8\times 1$}   & $61.00$\scalebox{0.7}{$\pm 00.17$}   & $57.60$\scalebox{0.7}{$\pm 00.48$}     & $61.15$\scalebox{0.7}{$\pm 00.19$}     \\ \hline
\multicolumn{1}{|c|}{$8\times 6\times 1$}   & $63.38$\scalebox{0.7}{$\pm 00.88$}   & $62.12$\scalebox{0.7}{$\pm 00.54$}     & $65.04$\scalebox{0.7}{$\pm 01.19$}     \\ \hline \hline
\multicolumn{1}{|c|}{$4\times 7\times 2$}   & $51.79$\scalebox{0.7}{$\pm 00.24$}   & $49.33$\scalebox{0.7}{$\pm 00.74$}     & $52.67$\scalebox{0.7}{$\pm 00.49$}     \\ \hline
\multicolumn{1}{|c|}{$7\times 4\times 2$}   & $60.17$\scalebox{0.7}{$\pm 00.12$}   & $56.56$\scalebox{0.7}{$\pm 00.35$}     & $61.50$\scalebox{0.7}{$\pm 00.47$}     \\ \hline
\multicolumn{1}{|c|}{$7\times 8\times 1$}   & $68.08$\scalebox{0.7}{$\pm 00.37$}   & $64.21$\scalebox{0.7}{$\pm 00.25$}     & $68.17$\scalebox{0.7}{$\pm 00.68$}     \\ \hline
\multicolumn{1}{|c|}{$8\times 7\times 1$}   & $68.75$\scalebox{0.7}{$\pm 00.82$}   & $64.77$\scalebox{0.7}{$\pm 00.56$}     & $68.63$\scalebox{0.7}{$\pm 00.33$}     \\ \hline \hline
\multicolumn{1}{|c|}{$6\times 6\times 2$}   & $60.40$\scalebox{0.7}{$\pm 01.54$}   & $58.00$\scalebox{0.7}{$\pm 00.23$}     & $62.73$\scalebox{0.7}{$\pm 01.31$}     \\ \hline
\multicolumn{1}{|c|}{$8\times 9\times 1$}   & $73.06$\scalebox{0.7}{$\pm 00.19$}   & $69.44$\scalebox{0.7}{$\pm 00.17$}     & $73.04$\scalebox{0.7}{$\pm 00.37$}     \\ \hline
\multicolumn{1}{|c|}{$9\times 8\times 1$}   & $74.40$\scalebox{0.7}{$\pm 00.52$}   & $71.31$\scalebox{0.7}{$\pm 00.39$}     & $74.15$\scalebox{0.7}{$\pm 00.19$}     \\ \hline
\multicolumn{1}{|c|}{$12\times 6\times 1$}  & $74.48$\scalebox{0.7}{$\pm 00.23$}   & $72.23$\scalebox{0.7}{$\pm 00.29$}     & $74.48$\scalebox{0.7}{$\pm 00.15$}     \\ \hline
\multicolumn{1}{|c|}{$9\times 4\times 2$}   & $66.40$\scalebox{0.7}{$\pm 00.19$}   & $64.44$\scalebox{0.7}{$\pm 00.35$}     & $67.67$\scalebox{0.7}{$\pm 01.04$}     \\ \hline \hline
\multicolumn{1}{|c|}{$10\times 10\times 1$} & $80.00$\scalebox{0.7}{$\pm 00.21$}   & $76.75$\scalebox{0.7}{$\pm 00.29$}     & $79.44$\scalebox{0.7}{$\pm 00.96$}     \\ \hline
\multicolumn{1}{|c|}{$10\times 5\times 2$}  & $72.92$\scalebox{0.7}{$\pm 00.25$}   & $70.00$\scalebox{0.7}{$\pm 00.41$}     & $71.12$\scalebox{0.7}{$\pm 00.38$}     \\ \hline \hline
\multicolumn{1}{|c|}{average}               & $65.16$                           & $62.25$      & $65.57$         \\ \hline \hline
\multicolumn{1}{|c|}{$\sum$epochs}               & $3150$                       & $210$       & $660$           \\ \hline
\end{tabular}%
}
\end{table}

\section{Conclusions}\label{conclusion}
In this paper, we proposed a novel training method and practical implementation for Multilinear Compressive Learning models that are capable of compressive signal acquisition and prediction with an adaptive compression rate. By enabling such a functionality in a remote compressive learning system, i.e. an adjustable degree of signal fidelity, the amount of data transmitted for each sample can be adjusted according to the conditions in the network. This can result in major improvements in the information content throughput of a remote sensing and learning application. Empirical evaluation of the proposed training approach showed that one can save a significant amount of computations during the training phase. Furthermore, when the server  can handle multiple model instances, one can achieve performances comparable to the standard setup while still having the adaptive compression rate feature on the sensing devices and lowering the number of training computations significantly. 

\section{Acknowledgement}
This project has received funding from the European Union’s Horizon 2020 research and innovation programme under grant agreement No 871449 (OpenDR). This publication reflects the authors’ views only. The European Commission is not responsible for any use that may be made of the information it contains.

The authors wish to acknowledge CSC – IT Center for Science, Finland, for computational resources.

\bibliography{reference}
\bibliographystyle{ieeetr}

\end{document}